\documentclass[11pt,twocolumn]{article}

\usepackage{graphicx}
\usepackage{algorithm}
\usepackage{algorithmic}
\usepackage{amsmath}
\usepackage{tikz}
\usepackage{pgfplots}
\usepackage{pgfplotstable}
\usepackage{filecontents}
\usetikzlibrary{calc, shapes.callouts}
\usetikzlibrary{plotmarks}
\usetikzlibrary{patterns}
\usetikzlibrary{mindmap,backgrounds}
\usetikzlibrary{positioning,shapes,shadows,arrows}
\usetikzlibrary{fit}
\usepackage{authblk}

\DeclareMathOperator*{\argmax}{arg\,max}

\begin{document}

\title{Probabilistic Cascading for Large Scale Hierarchical Classification}

\author[1,2]{Aris Kosmopoulos}
\author[1]{Georgios Paliouras}
\author[2,3]{Ion Androutsopoulos}

\affil[1]{Institute of Informatics and Telecommunications, National Center for Scientific Research ``Demokritos'', Athens, Greece}
\affil[2]{Department of Informatics, Athens University of Economics and Business, Greece}
\affil[3]{Institute for the Management of Information Systems (Digital Curation Unit) and Institute for Language and Speech Processing, Research Center ``Athena'', Athens, Greece}

\maketitle

\begin{abstract}
Hierarchies are frequently used for the organization of objects. Given a hierarchy of classes, 
two main approaches are used, to automatically classify new instances: flat classification and cascade classification.
Flat classification ignores the hierarchy, while cascade classification greedily traverses the hierarchy from the root to the predicted leaf. 
In this paper we propose a new approach, which extends cascade classification to predict the right leaf by estimating the probability of each root-to-leaf path. 
We provide experimental results which indicate that, using the same classification algorithm, one can achieve better results with our approach, compared 
to the traditional flat and cascade classifications.

\end{abstract}

\section{Introduction}

Machine learning is often used to estimate classification models for a set of predefined categories. Most of the times, these categories are assumed to be independent.
When independence cannot be assumed we may either construct artificial hierarchies (hierarchical clustering), or classify new instances onto a hierarchy 
that is given, typically representing is-a relations. 

In this paper we study cases where the hierarchy is already provided. Furthermore, the hierarchy is a tree and the classification nodes are always the 
leaves of the hierarchy. We also assume that each instance belongs to only one category (single-label classification).

Many researchers approach hierarchical classification problems \cite{Brouard12,TsoumakasLMV13} using flat classification, i.e. ignoring the hierarchy. 

Hierarchical classification approaches, typically divide the problem into smaller ones, usually one classification for each node of the hierarchy. 
For each of these problems fewer features and instances are required to train a good classifier, compared to the respective flat approaches. 
This can be very important, especially in cases of large scale classification, where the number of categories and instances can increase to thousands 
and millions respectively. In such cases, a hierarchical approach would require much fewer resources than a flat one.

The main issue in hierarchical classification is to combine the decisions of the node-specific classifiers appropriately, in order to predict a category for an instance. 
The most common approach is that of cascade classification. In this case, we start at the root of the hierarchy and greedily select the most probable 
descendant. This continues until we reach a leaf, which is chosen as the predicted node. 
The main disadvantage of this approach is that any mistake done during the descent deterministically leads to the wrong final decision. 
Therefore the cascade is very sensitive to the quality of the inner node classifiers. 
In this paper we propose a new approach, which is as fast as cascade regarding training but leads to better results compared to cascade and flat classification, 
using the same classification algorithms.

In the next Section we present the related work, while in Section 3 we introduce our approach. Section 4 discusses our experimental results. 
Finally, Section 5 concludes and points to future work.

\section{Related Work}

Although hierarchical classification has many advantages, typically researchers resort to mildly hierarchical or even flat approaches \cite{kosmopoulos2010}.
One reason for this is that flat classification is well studied, so it is easier to transfer methods from this field. 
On the other hand on large scale problems, the flat use of traditional classifiers, such as SVMs, is often prohibitively expensive computationally \cite{Liu:2005}. 

Early work in hierarchical classification focused on approaches such as shrinkage \cite{McCallumRMN98} and hierarchical mixture models \cite{ToutanovaCPH01}.
Unfortunately most of these approaches cannot be applied to large scale problems, at least in the form described in the original papers. 
New methods based on similar ideas, such as that of latent concepts \cite{QiuHLZ11}, continue to appear in the literature, taking also into account 
scalability issues. But still most of the proposed methods are tested on rather small datasets with small hierarchies.

Mildly hierarchical approaches, typically make limited use of the hierarchy. Methods such as \cite{XueXYY08} use only some levels
of the hierarchy, flattening the rest. Other approaches such as \cite{BabbarPGA13}, alter the initial hierarchy before performing cascading in order 
to minimize errors at the upper levels of the hierarchy.

\section{Probabilistic Cascading}

\begin{figure}[h]
\begin{center}
 \begin{tikzpicture}[-,>=stealth',shorten >=1pt,auto, node distance=3cm,
  thick,main node/.style={circle,fill=blue!20,draw,font=\sffamily\normalsize\bfseries, minimum size=14mm, scale=0.6}]

\node[main node] (1) {Root};
\node[main node] (2) [below left of=1] {Arts};
\node[main node] (3) [below right of=1] {Health};
\node[main node] (4) [below left of=2] {Music};
\node[main node] (5) [below right of=2] {Dance};
\node[main node] (6) [below right of=3] {Medicine};
\node[main node] (7) [below of=3] {Fitness};

\path[every node/.style={font=\sffamily\small}]
    (1) edge node [left] {} (2)
	edge node [right] {} (3)
    (2) edge node [left] {} (4)
	edge node [right] {} (5)
    (3) edge node [right] {} (6)
        edge node [right] {} (7);

\end{tikzpicture}
\caption{Tree hierarchy example.}
\label{treeExample}
\end{center}
\end{figure}
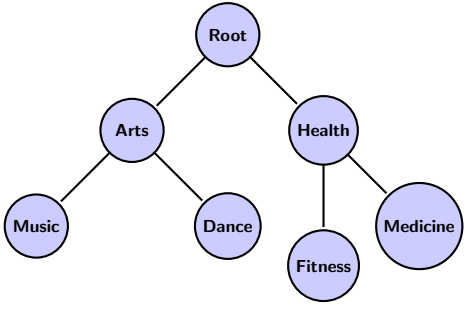

In our method following the cascading approach, we train one binary classifier for each node of the hierarchy. 
For example, using the hierarchy of Figure \ref{treeExample}
we would train one classifier for each of the nodes Arts, Health, Music, Dance, Fitness and Medicine. 
The binary classifier of a node N is trained using as positive examples the instances belonging to the leaf descendants of N and as negative examples the
instances of its siblings. For example, the binary classifier of node Music would use all instances belonging to Music as positive  
examples and all instances belonging to Dance as negative examples. Similarly for the binary classifier of node Arts, all instances belonging to Music and Dance would be positive examples,
while all instances belonging to Fitness and Medicine would be negative.

These binary classifiers require fewer resources to be trained compared to flat ones. 
They can also be more accurate, since they aim to distinguish between fewer categories. 
For example, if we have 10,000 leaves, each binary classifier would need to separate one class from 9.999 others. 
In the case of cascading,it would only need to separate between the sibling categories. 
Such classifiers would also require fewer features to train on, an important characteristic if we consider large datasets.

The main disadvantage of cascading is that any mistake is carried over. For example if an instance belonging to category of Music, 
gets a higher probability by the classifier of Health than that of Arts, is classified wrongly, without taking into consideration the 
classifiers of Music and Dance. In contrast, our method computes the probability of each root-to-leaf path for a testing instance and we classify
it to the most probable path, which we call $P_{path}$. As an example, the probability of an instance $d$ belonging to Music:

\begin{gather}
P(Music|d) = \frac{P(Arts|d) P(Music|Arts,d)}{P(Arts|Music,d)}
\end{gather}
but  since $P(Arts|Music,d)=1$:
\begin{gather}
P(Music|d) = P(Arts|d) P(Music|Arts,d) 
\end{gather}
Similarly:
\begin{gather}
P(Arts|d) = \frac{P(Root|d) P(Arts|Root,d)}{P(Root|Arts,d)}
\end{gather}
but  since $P(Music|Root,d)=1$ and $P(Root|d)$ = 1:
\begin{gather}
P(Arts|d) = P(Arts|Root,d)
\end{gather}
By combining (2) and (4) we get:
\begin{gather}
P(Music|d) =  P(Arts|Root,d)P(Music|Arts,d) 
\end{gather}

These conditional probabilities are in fact the ones computed by the binary classifiers of each node.
So given a document $d$, a leaf $C$ and a set $S$ of all the ancestors of $C$: 
\begin{gather}
 P(C|d) = \prod_{i=1}^{|S|}{P(S_{i}|Ancestor(S_{i}),d)}
\end{gather}

and we define $P_{path}$ as the:

\begin{gather}
P_{path}(d) = \argmax _{C}P(C|d)
\end{gather}

Let's get back to our initial example where document $d$ belonged to Music. Lets assume that we have the following probabilities:
\begin{itemize}
 \item $P(Arts|Root,d) = 0.2$
 \item $P(Health|Root,d) = 0.21$
 \item $P(Music|Arts,d) = 0.9$
 \item $P(Dance|Arts,d) = 0.6$
 \item $P(Fitness|Health,d) = 0.1$
 \item $P(Medicine|Health,d) = 0.2$
\end{itemize}

If we used standard cascading, document $d$ would be classified to category Medicine.
Using $P_{path}$ we get:
\begin{itemize}
 \item $P(Music|d) = 0.18$
 \item $P(Dance|d) = 0.12$
 \item $P(Fitness|d) = 0.021$
 \item $P(Medicine|d) = 0.042$
\end{itemize}

and $P_{path}$ would assign $d$ to class Music. The cost that we have to pay, compared to standard cascading, is that we have to compute all the
$P(C|d)$, in order to select the one with the highest probability. 
\section{Experimental results}

In order to compare our approach against flat and cascade classification, we used the Task 1 dataset of the first Large Scale Hierarchical Text Classification Challenge (LSHTC1).\footnote {http://lshtc.iit.demokritos.gr/node/1}
This dataset contains 93,505 instances (split into train and validation files), composed of 55,765 distinct features and belonging to 12,294 categories. Classification is only allowed to the leaves of the hierarchy, which is a tree. 
Each instance belongs to only one category.
The testing instances are 34,880 and the results are evaluated using the evaluation measures of the challenge (an Oracle is provided by the organizers) 
which are the following:
\begin{itemize}
 \item Accuracy
 \item Macro F-measure
 \item Macro Precision
 \item Macro Recall
 \item Tree Induced Error
\end{itemize}

As a classifier we used a L2 Regularized Logistic Regression with the regularization parameter C set to 1 (usually the default value). We also conducted 
experiments with other regularization methods and other values of C, but the results were similar. All the experiments were conducted using TF/IDF instead 
TF features, as our experiments indicated better performance with this feature set. 

The goal of our experiments was to illustrate that the proposed method can improve the results of flat and cascade classification, using the same algorithm,
L2 Regularized Logistic Regression in this case. Further experimentation and engineering could make the method competitive to the best-performing systems, in the 
challenge. However, we consider this exercise beyond the scope of the paper.

For flat classification, we trained one binary classifier (one versus all) for each leaf. We then assigned each testing document to the class  with the highest probability. 
For cascade classification we trained a binary classifier for each node of the hierarchy. We used as positive examples, all the instances belonging to all the descendant leaves of the node and 
as negative, all the descendant leaves of its siblings. 
This results in more classifiers than the for flat classification, but each of these classifiers was much easier to train, since it was 
trained on fewer instances.

In table \ref{tbl:results_tf}, we present the results of each approach, for each evaluation measure. 
The main observation is that $P_{path}$ outperforms both Flat and Cascade. 
Another interesting result is, that Flat is the worst approach, according to Tree Induced Error. 
This is an indication that by ignoring the hierarchy (flat classification), the mistakes tend to be located further from the correct category in the hierarchy.
This is very important in hierarchical classification, since different mistakes carry different weight.
Misclassifying an instance to a sibling of the correct category is a smaller error than if it was classified to a category 5 nodes away. 
Flat evaluation measures, generally fail to capture this, so tree induced error, being the only hierarchical measure of the five that we use is more suitable for comparing
the three approaches.

Given that our hierarchy is a tree and each instance belongs to only a single class, there is no need to take into account more complex
hierarchical evaluation measures and tree induced error is sufficient for safe conclusions.

\begin{table}[h]
\centering
\scalebox {0.8}{
\begin{tabular}{|l | c c c|}
\hline
Evaluation Measure & Flat & Cascade & $P_{path}$\\
\hline
Accuracy & 0.405 & 0.404 & \textbf{0.431} \\
Macro F-measure & 0.256 & 0.278 & \textbf{0.294} \\
Macro Precision & 0.254 & 0.269 & \textbf{0.287} \\
Macro Recall & \textbf{0.302} & 0.289 & \textbf{0.302} \\
Tree Induced Error & 3.874 & 3.609 & \textbf{3.437} \\
\hline
\end{tabular}
}
\caption{Results for each approach per evaluation measure,  using TF/IDF features. With bold we mark the best performing approach, given each evaluation measure.}
\label{tbl:results_tf}
\end{table}

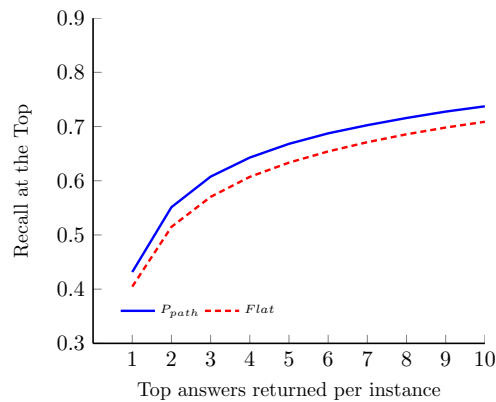
\begin{figure}[h]
\begin{center}
\begin{tikzpicture}[scale=0.76]
    \begin{axis}[
    legend columns=2,
    legend style={ at={(0.05,0.1)},anchor=west,draw=none},
    xlabel={\small Top answers returned per instance},
    ylabel={\small Recall at the Top},
    ymin=0.3, ymax=0.9, xmin=0, xmax=10, 
    ytick={0.3, 0.4, 0.5, 0.6, 0.7, 0.8, 0.9},
    xtick={1, 2, 3, 4, 5, 6, 7, 8, 9, 10},
    thick,
    axis x line*=bottom,
    axis y line*=left]

     \pgfplotstableread{
     Acc	K
      0.431565	1
      0.551233	2
      0.607597	3
      0.642775	4
      0.668091	5
      0.687443	6
      0.702523	7
      0.715768	8
      0.72758	9
      0.737328	10
     }\Ppath
     
     \pgfplotstableread{
     Acc	K
     0.404633	1
     0.514865	2
     0.570282	3
     0.60715	4
     0.633468	5
     0.654081	6
     0.671139	7
     0.685731	8
     0.698145	9
     0.708839	10
     }\Flat
    
     \addplot[color=blue,very thick] table[y=Acc,x=K] {\Ppath};
 	\addlegendentry{\tiny $P_{path}$}
    \addplot[color=red,very thick, densely dashed] table[y=Acc,x=K] {\Flat};
 	\addlegendentry{\tiny $Flat$}
    \end{axis}    
\end{tikzpicture}
\label{topK}
\caption{Recall at the top K answers of $P_{path}$ and Flat classification for various values of $K$.}
\end{center}
\end{figure}

Both $P_{path}$ and Flat classification produce a probability for each leaf and the highest one is returned as the predicted category. 
But what if we evaluated the list of categories, ranked according o their probability? In order to obtain such an assessment in Figure \ref{topK} 
we calculate the recall for the $K$ most-probable categories, with $K$ ranging from 1 to 10. 
As expected the probability of success increases rapidly with $K$. 
This is very important, for a realistic semi-automated classification scenario, where a human annotator selects the correct label between thousands of categories.
Such a system would allow the annotator to select only between five or ten suggestions.
The second observation is that for all values of $K$, $P_{path}$ performs better than the Flat one.

Regarding the scalability of the approaches, during the two cascading approaches (standard and $P_{path}$) require fewer resources than the flat classifiers.
During classification, $P_{path}$ is slower than Cascade, since it takes into account all the root-to-leaf paths,  
and is similar to the cost of Flat classification.

\section{Conclusions}

In this paper we present the $P_{path}$ method for hierarchical classification. $P_{path}$ addresses the disadvantages of traditional flat and cascade classification. 
Flat classification can be very computational demanding in large scale problems and also ignores completely the hierarchy information which can be exploited for better results.
Standard cascading on the other hand is much more computational efficient, but suffers from the problem of early misclassification at the top levels of the hierarchy.

Our approach has the same training computational complexity as the Cascade, while achieving better scores according to all the tested evaluation measures. 
However, it is slower during classification, having a complexity is similar to that of flat classification.

The version presented in this paper is designed for tree hierarchies. As a future work, we plan to extend the idea of $P_{path}$ to DAG hierarchies. 
Furthermore in this paper we focused on single-label classification. Although the idea of $P_{path}$ seems 
compatible with multi-label approaches, further experiments need to be conducted in this direction.

\bibliographystyle{plain}
\bibliography{ProbabilisticCascade}

\begin{thebibliography}{1}

\bibitem{BabbarPGA13}
Rohit Babbar, Ioannis Partalas, Eric Gaussier, and Massih-Reza Amini.
\newblock Maximum-margin framework for training data synchronization in
  large-scale hierarchical classification.
\newblock In {\em Neural Information Processing}, pages 336--343, 2013.

\bibitem{Brouard12}
C.~Brouard.
\newblock Document classification by computing an echo in a very simple neural
  network.
\newblock In {\em ICTAI}, pages 735--741, 2012.

\bibitem{kosmopoulos2010}
A.~Kosmopoulos, {\'E}.~Gaussier, G.~Paliouras, and S.~Aseervatham.
\newblock The ecir 2010 large scale hierarchical classification workshop.
\newblock {\em SIGIR Forum}, 44(1):23--32, 2010.

\bibitem{Liu:2005}
T.~Liu, Y.~Yang, H.~Wan, H.~Zeng, Z.~Chen, and W.~Ma.
\newblock Support vector machines classification with a very large-scale
  taxonomy.
\newblock {\em SIGKDD Explor. Newsl.}, 7(1):36--43, June 2005.

\bibitem{McCallumRMN98}
A.~McCallum, R.~Rosenfeld, T.~M. Mitchell, and A.~Y. Ng.
\newblock Improving text classification by shrinkage in a hierarchy of classes.
\newblock In {\em ICML}, pages 359--367, 1998.

\bibitem{QiuHLZ11}
X.~Qiu, X.~Huang, Z.~Liu, and J.~Zhou.
\newblock Hierarchical text classification with latent concepts.
\newblock In {\em ACL (Short Papers)}, pages 598--602, 2011.

\bibitem{ToutanovaCPH01}
K.~Toutanova, F.~Chen, K.~Popat, and T.~Hofmann.
\newblock Text classification in a hierarchical mixture model for small
  training sets.
\newblock In {\em CIKM}, pages 105--112, 2001.

\bibitem{TsoumakasLMV13}
G.~Tsoumakas, M.~Laliotis, N.~Markantonatos, and I.~P. Vlahavas.
\newblock Large-scale semantic indexing of biomedical publications.
\newblock In {\em BioASQ@CLEF}, 2013.

\bibitem{XueXYY08}
G.~Xue, D.~Xing, Q.~Yang, and Y.~Yu.
\newblock Deep classification in large-scale text hierarchies.
\newblock In {\em SIGIR}, pages 619--626, 2008.

\end{thebibliography}
\end{document}